\documentclass[conference]{IEEEtran}

\usepackage{amsmath}
\usepackage{amssymb}
\usepackage{graphicx}      
\usepackage{subfigure}
\graphicspath{{figures/}}

\begin{document}

\title{A semantic-aided particle filter approach for AUV localization}

\author{\IEEEauthorblockN{
Francesco Maurelli, Szymon Krupi\a'nski}
\IEEEauthorblockA{
Marine Systems and Robotics\\
Jacobs University Bremen\\
Bremen, DE\\
Email: f.maurelli@ieee.org}}


%


\maketitle

\begin{abstract}
This paper presents a novel approach to AUV localization, based on a semantic-aided particle filter. Particle filters have been used successfully for robotics localization since many years. Most of the approaches are however based on geometric measurements and geometric information and simulations. In the past years more and more efforts from research goes towards cognitive robotics and the marine domain is not exception. Moving from signal to symbol becomes therefore paramount for more complex applications. This paper presents a contribution in the well-known area of underwater localization, incorporating semantic information. An extension to the standard particle filter approach is presented, based on semantic information of the environment. A comparison with the geometric approach shows the advantages of a semantic layer to successfully perform self-localization.
\end{abstract}


\section{Introduction} 

Robot localization has been researched by scientists from the very beginning of robotics. Knowing its position and orientation is an essential task for an autonomous system in many circumstances.
Very often, it is a prerequisite to perform more complex tasks.
The marine domain is no exception. In most - if not all - real life scenarios, the robot location is an essential information.
Marine robots are increasingly used to perform a great variety of tasks, ranging from oil\&gas applications to defense, from marine biology to underwater archaeology. In all of these scenarios, the robot location is fundamental.
There are several challenges to perform underwater localization. The lack of GPS signal is the most evident one. In order to overcome this, various acoustic-based solution can be employed, like for example Long BaseLine (LBL) acoustic positioning system. This requires to deploy acoustic transponders as aid for the vehicle, which can compute its location with a triangulation from the data received by the transponders. The drawback of this technique is however the need to actively deploy external transponders, which cause additional cost, time and logistic challenges. Additionally, the GPS location at the drop-in point might not be the same than the GPS on the seabed, especially in deep sea, with strong currents.
Many offshore infrastructures are located in environments which fall into this category.
For this reason, several techniques have been used in the past years to allow an underwater vehicle to determine its location based entirely on the on-board sensor suite.
Geometric approaches were developed based on distance sensors like sonar, in parallel with geometric approaches developed in land robotics, based on laser scanners.
In recent years there has been a substantial interest from the research community to explore semantic aspects of knowledge representation, and its influence in the vehicle's tasks. Generally speaking, robots still lack the high-level abstraction capability typical of humans. This is a complex problem, as it aims to shift the paradigm from sensor processing into a more organized, long-term knowledge structure in robotics systems, with possibility of augment, reasoning and learn.
This paper represents a step in this direction, with the of use of semantic information in processes traditionally covered only by geometric approaches.
The paper is organized as follows:\\
- Section II will present the state-of-art and the related work\\
- Section III will present the technical approach that was undertaken\\
- Section IV will present the results\\
- Section V will critically discuss limitations and improvements of the proposed approach

\section{Related Work}
AUV localisation is a problem which has been studied from the very early stages of marine autonomy. Already in 1990, Rigaud \textit{et al.} proposed a system based on sensor fusion to correct dead reckoning~\cite{AUVloc1990}, using a discrete stochastic-accumulation method. The results are shown postprocessing sonar data from a pool.
Particle filters, introduced by Gordon in 1993 \cite{Gordon1993}, have been an effective way of representing the evolution of a probability distribution by a set of samples. They have now become one of the \textit{standard} filtering techniques in robotics. Their adoption in the underwater domain was however slower than in other robotics area, due to the more limiting conditions, such as reduced computational power in the pressure hull(s). Kondo \textit{et al.} proposed a real-time localization method for navigation based on particle filters that can fuse multisensor data - image and acoustic-based profiling systems \cite{Kondo2004}. Karlsson proposed a particle filter
approach for AUV navigation, with a focus on mapping \cite{Karlsson2003}. Silver \textit{et al.} presented
particle filter merged with scan matching techniques \cite{Silver2004}. Used with an approximation of the likelihood of sensor readings,
based on nearest neighbor distances, particle filters are able to
approximate the probability distribution over possible poses. Wirth \textit{et al.} addressed the use of particle filters for visual
tracking of underwater elongated structures such as cables or
pipes \cite{Wirth2008}. Models for probabilistic tracking are obtained directly from real underwater image sequences.
Previous work from some of the authors include the application of particle filters and Kalman filters in a variety of underwater settings~\cite{Maurelli2008,Maurelli2009,Petillot2010}.
Those approaches are mainly geometric, namely the goal is to determine the robot's pose (location and orientation), based on geometric measurements (or transformed to geometric measurements).
In the recent years, an emerging area in robotics has been dealing with more cognitive aspects linked to the world representation, which extend the geometric-only aspects. Semantic representation has become an active field of research, shifting the research focus from \textit{signal} to \textit{symbol}.
Semantic information have been used in marine robotics in various scenarios, with the goal of improving situation awareness \cite{Miguelanez:Semantic:09}. Projects focused on persistent autonomy in the marine domain has addressed semantic representation, linking it to robust long-term planning and dynamic replanning \cite{pandora2012T, pandora2016}. In this work we aim to build on top of the results in knowledge representation and in localization, to address semantic localization, an area which is still very challenging, especially in the marine domain.
Lim \& Sinha proposed an approach to detect semantic features to use  in a Kalman filter \cite{Sinha2012}. The proposed 2D-to-3D matching approach for recognizing 3D points in the map does not require the construction of an explicit semantic map. Rather semantic information can be associated with the 3D points in the reconstruction process and
can be retrieved via recognition during online localization.
Other approaches focused on feature detection, focusing on the 3D Point Cloud and the PCL Library \cite{Martinez-Gomez2016}, or on feature-based scene descriptors from images, using dynamic Bayesian networks \cite{rubio2014dynamic}.
Other approaches aim to determine the semantic place the robot is. For example Villena \textit{et al.} proposed a multimodal HRI approach, with a finite set of places where the robot could be \cite{Villena2015}. This approach to the problem of semantic localization however cut off completely the geometric dimension of the localization problem, whereas the proposed work aims to use the aid of semantic knowledge to compute the robot pose in geometric terms.
Yi \textit{et al.} uses a particle filter approach exploiting semantic relations among objects to aid the localization process \cite{yi2009active}. This approach uses spatial relations among objects in the environment recognized in camera images.
The proposed approach in this paper is similar to the last approach as general structure of the localization approach (without the active component), though there are important differences in the sensor suite, in the way knowledge is represented and therefore in the computation of the likelihood function. No significant semantic approach for localization in the marine robotics domain is currently available in the related literature.

\section{Technical Approach}
A particle filter is a Bayes
filter that works by representing a probability distribution $p(x$
as a set of samples, as expressed in the following equation.
\begin{equation}
p(x) \approx \dfrac{1}{N} \sum\limits_{i} \delta_{x^{i}}(x)
\end{equation}
where $N$ represents the number
of samples, $x^{(i)}$ is the state of the sample $i$, $ \delta_{x^{i}}(x)$ is
the impulse function centered in $x^{i}$. The more dense the
samples $x^{i}$ in a region, the higher is the probability that the
current state falls within that region. In principle, in order to
maintain a sample (particle) representation of the system state
distribution over the time $t$, the samples $x^{i}_t$ should be created
from the probability distribution of the current state, given the
observation history $z_{0:t}: p(x_t | z_{0:t})$. Such a distribution is in
general not available in a form suitable for sampling. However,
the importance sampling principle ensures that if:
\begin{itemize}

\item we are able to evaluate pointwise and to draw samples
from an arbitrarily chosen importance function $\pi(x_t | z_{0:t})$,
such that $(p(x_t | z_{0:t}) > 0) ) \Rightarrow (\pi(x_t | z_{0:t}) > 0)$, and
\item we are able to evaluate pointwise $p(x_t | z_{0:t})$,
\end{itemize}
then it is possible to recover a sampled approximation of
$p(x_t | z_{0:t})$ as outlined in the following equation:
\begin{equation}
p(x_t | z_{0:t}) \propto \sum\limits_{i} w^{(i)} \delta_{x_{t}}^{(i)}
\end{equation}

where $x_t^{(i)}$ are samples drawn from $\pi(x_t | z_{0:t})$ and
$w^{(i)}_t~=~\dfrac{p(x^{(i)}_t | z_{0:t})}{\pi(x^{(i)}_t | z_{0:t})}$ is the importance
weight related to the $i^{th}$ sample that takes into account
the mismatch among the target distribution $p(x_t | z_{0:t})$ and
the importance function. 
One of the most common particle filtering algorithms is
the Sampling Importance Resampling (SIR) filter. A
SIR filter incrementally processes the observations $z_t$ and the
commands $u_t$ (process evolution), by updating a set of samples
representing the estimated distribution $p(x_t | z_{1:t}, u_{0:t})$. This is
done by performing the following three steps:
\begin{itemize}

\item \textit{Sampling}: The next generation of particles $x^{(i)}_t$ is obtained
by the previous generation $x^{(i)}_{t-1}$, by sampling from
a proposal distribution $\pi(x_t | z_{1:t}, u_{0:t})$.
\item \textit{Importance Weighting}: An individual importance weight
$w^{(i)}$ is assigned to each particle, according to the following equation:

\begin{equation}
w^{(i)} = \dfrac{p(x_t | z_{1:t}, u_{0:t})}	
{\pi(x_t | z_{1:t}, u_{0:t})}
\end{equation}

The weights $w^{(i)}$ account for the fact that the proposal
distribution in general is not equal to the true distribution
of the successor states.
\item \textit{Resampling}: Particles with a low importance weight $w$ are
typically replaced by samples with a high weight. This
step is necessary since only a finite number of particles
are used to approximate a continuous distribution. Furthermore,
resampling allows to apply a particle filter in
situations in which the true distribution differs from the
proposed one.

\end{itemize}

Moving from a pure geometric approach to a semantic-aided one, the main function that needs some modification is the \textit{Importance Weighting}. In a pure geometric approach, the weight is calculated comparing arrays of distances, given that every observation from the robot can be translated into the geometry of the surrounding landscape. In the chosen semantic-aided approach, each observation is formed by \textit{q} objects, with a relative pose with respect to the robot pose:
\begin{equation}
z^k: Object_1(\rho_1,\theta_1), Object_2(\rho_2,\theta_2), ... , Object_q(\rho_q,\theta_q)
\end{equation}

The calculation of the weight of a particle means to evaluate how close two sets of observations $z^r$ and $z^k$ are.
$z^r$ can be expressed in the same way than the above mentioned $z^k$. assuming $l$ objects observed:
\begin{equation}
z^r: Object_1(\rho_1,\theta_1), Object_2(\rho_2,\theta_2), ... , Object_l(\rho_l,\theta_l)
\end{equation}
A mixture of two families of Gaussians for the $m$ objects in common between the two observations is employed to calculate a first estimation of the weight. All Gaussians are centered in 0, with the value calculated at the points: $\{\rho_1^r-\rho_1^k, \theta_1^r-\theta_1^k, ..., \rho_m^r-\rho_m^k, \theta_m^r-\theta_m^k\}$

The calculated value is then adjusted taking into consideration the $d$ objects present in one observation, but not in the other one, with $d = q+l-2m$.

\section{Simulated Setup and Results}
\begin{figure}
	\begin{center}
\includegraphics[width=0.4\textwidth]{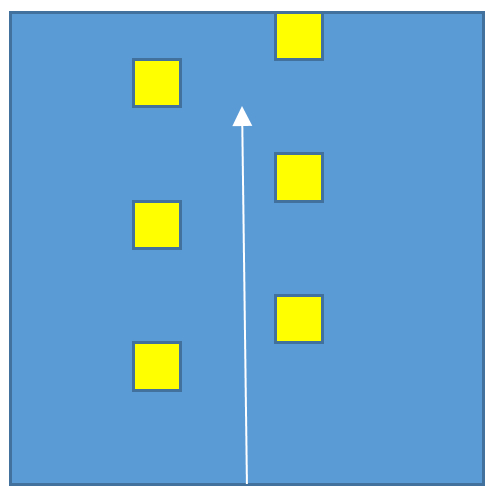}
	\end{center}
	\caption{The simulated scenario: the AUV moves close to man-made structures and use them to localize itself. The size of the environemnt is is 50x50 meters, with each block being a square 5x5 meters. The white line represents the vehicle's intended trajectory.}
	\label{map}
\end{figure}

\begin{figure}
	\begin{center}
\includegraphics[width=0.5\textwidth]{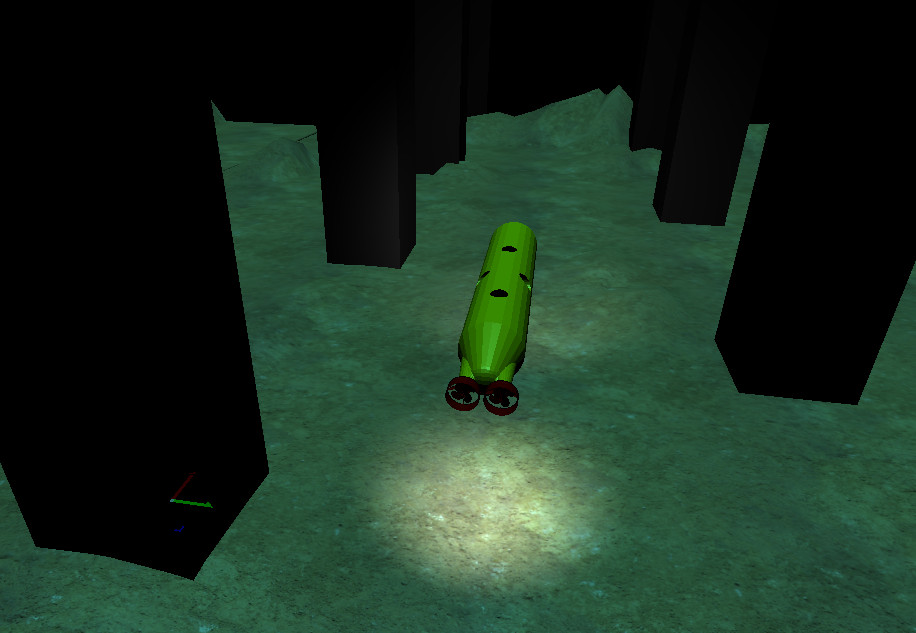}
	\end{center}
	\caption{A screenshot of the AUV simulator based on Morse.}
	\label{morse}
\end{figure}

The developed algorithm was integrated in the ROS-based AUV simulator used at Jacobs University. This has the advantage to use the same architecture that runs in the real AUV,
with simulated sensors and dynamics.
The environment prepared to evaluate the proposed semantic-aided localization approach is a 50x50m area, with the AUV moving close to various man-made structures, as shown in Figure~\ref{map}.
A screenshot of the simulator, based on Morse~\cite{Morse}, can be seen in Figure~\ref{morse}.

\begin{figure*}
	\begin{center}
\includegraphics[width=0.4\textwidth]{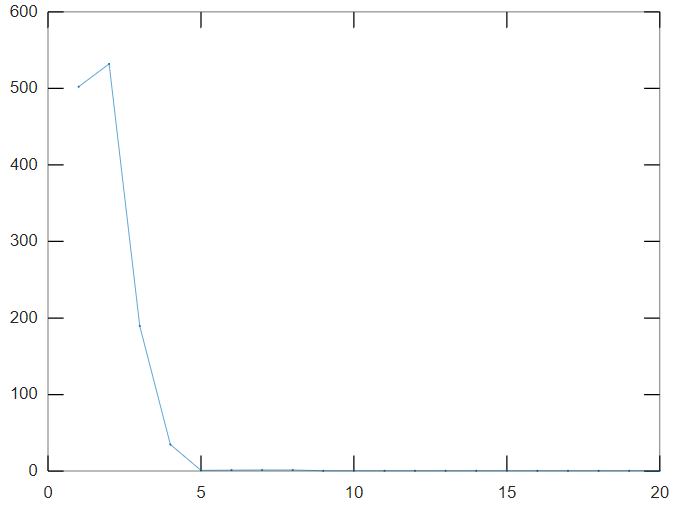}
\includegraphics[width=0.4\textwidth]{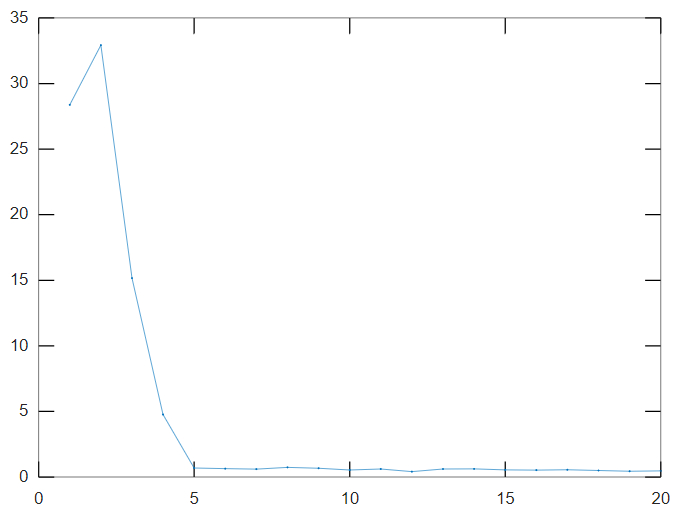}
\includegraphics[width=0.4\textwidth]{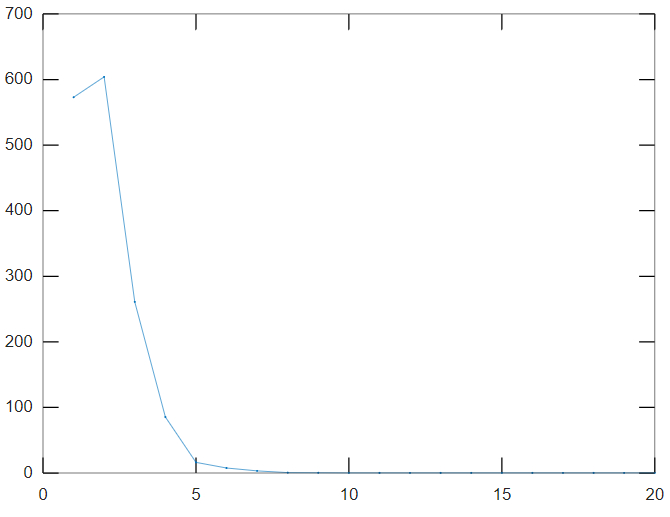}
\includegraphics[width=0.4\textwidth]{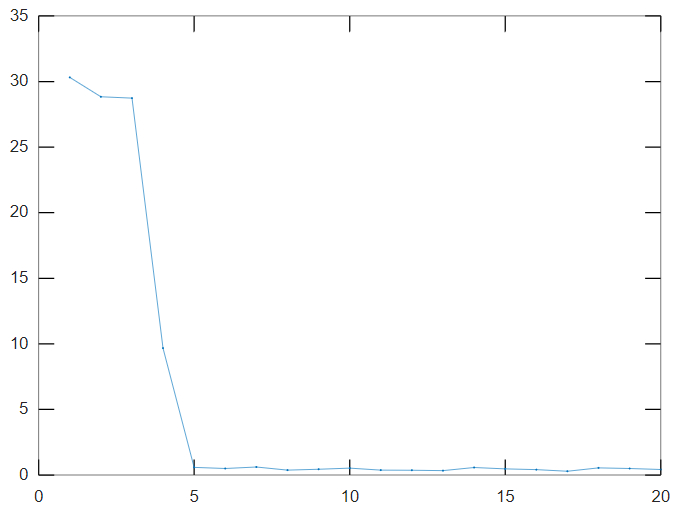}
	\end{center}
	\caption{Plots of the average particle variance (left) and the average error (right) in the localization, over 100 runs. A comparison among the geometrical approach (bottom) and the semantic approach (top) shows slightly better performances for the semantic-aided one. On the other hand, the time needed for semantic-aid localization is just a fraction than the one needed by the pure geometric one, as outlined in Table~\ref{time}. On the horizontal axes timestamps are represented, while the vertical axes represent the variance ($m^2$, left), and the error ($m$, right).}
	\label{stats}
\end{figure*}

Tests were run with the standard geometric approach and with the proposed semantic-aided one. 100 tests were run for each of the approach, and the results are shown in Figure~\ref{stats} and Table~\ref{time}. In the specific scenario, the average time to convergence, as well as the error and the particle variance are comparable in the two approaches. The proposed semantic-aided approach shows some improvements, but both results are in the same range. The real difference is however in the average time needed for the execution, with the semantic-aided approach being approximately six times faster than the classical geometrical one. This is particularly relevant in the underwater domain, where computational power is still an open issue, and it is not scalable in the same way than for land robotics.
A more efficient process is therefore pivotal for the actual use of the proposed approach in the field.

\begin{table}
\begin{center}
  \begin{tabular}{| c | c |}
    \hline
    \hline
\multicolumn{2}{ |c| }{Avg. time (ns) - 100 runs} \\
\hline
    Semantic-aided & Geometric \\ \hline
    1171934146 & 6977878010  \\ \hline
    \hline
  \end{tabular}
\end{center}
\caption{Average time in nanoseconds over 20 runs. Semantic-aided approach for localisation can have a huge result, with a 1:6 ratio as for time needed to execute the program.}
\label{time}
\end{table}

\section{Discussion and Conclusions}
This paper has presented a semantic-aided approach for AUV localization. Classical techniques are based on geometrical information only, whilst the proposed approach takes into consideration the transformation from signal to symbol and work with objects and their relative position to the robot.
The preliminary results shown in this paper are very encouraging, showing a very efficient process, about six times faster than its geometric counterpart.
One of the most demanding operation in any particle filter technique is the simulation of the observation from each particle. Having a semantic map stored in the vehicle knowledge base significantly reduce the complexity of this operation.
Simulating distances with respect to random surfaces would require computational demanding ray tracing algorithms, whilst the complexity of generating a semantic observation, given the robot pose and the semantic map, is significantly lower.
For the simulated set-up, ray-tracing techniques were even avoided, but analytic geometric formulas have been employed. These however can only be used in specific environments, like the one simulated. For more complex environments, ray tracing is the norm. Therefore, the efficiency shown by the semantic-aided approach can also be greater than the factor calculated, based on the way geometric observations are simulated.
On the other hand, there is an important step which was not in the scope of this paper, but it enters into play when addressing the overall system. This is the object recognition process, moving from the signals from the sensors to symbolic representation. This can be computational intensive, depending on the technique used and on the surrounding environment. This of course needs to be considered when talking about computational efficiency, though there are two important considerations: the first is merely linked on the times this operation would need to be invoked, and that is once per observation. This may well be worth in order to avoid using expensive ray tracing algorithms for thousands of particles. The second consideration is that there is an increasing effort from the research community in autonomous systems to include semantic aspects, build a semantic map, use a semantic map connected with the planning or the learning system. If a vehicle is already building and maintaining a semantic map, then using it for localization would not require additional efforts and sensor processing techniques than the ones already in place.
There are several future directions that we are interested in pursuing. First we will show a validation of this approach with real data gathered in previous missions. Due to various constraints, those results are not yet available for publication.
We are also interested in working with probabilistic semantic maps, rather than deterministic, as shown in this paper.
Finally, active techniques - already developed by some of the authors with a geometric setting - would benefit from a similar semantic-aided approach presented in this paper.

\section*{Acknowledgment}
The research leading to these results has received funding from the European Union Horizon2020 Programme - Marie Sk\l{}odowska-Curie Action - under grant agreement No. 709136 TIC-AUV.

\bibliographystyle{unsrt}
\bibliography{final}             

\begin{thebibliography}{10}

\bibitem{AUVloc1990}
V.~Rigaud, L.~Marce, J.~L. Michel, and P.~Borot.
\newblock Sensor fusion for auv localization.
\newblock In {\em Symposium on Autonomous Underwater Vehicle Technology}, pages
  168--174, Jun 1990.

\bibitem{Gordon1993}
N.~J. Gordon, D.~J. Salmond, and A.~F.~M. Smith.
\newblock Novel approach to nonlinear/non-gaussian bayesian state estimation.
\newblock {\em IEE Proceedings F - Radar and Signal Processing},
  140(2):107--113, April 1993.

\bibitem{Kondo2004}
H.~Kondo, T.~Maki, T.~Ura, Y.~Nose, T.~Sakamaki, and M.~Inaishi.
\newblock Relative navigation of an auv using image-and-acoustic based
  profiling systems.
\newblock In {\em OCEANS '04. MTTS/IEEE TECHNO-OCEAN '04}, volume~3, pages
  1330--1335 Vol.3, Nov 2004.

\bibitem{Karlsson2003}
R.~Karlsson, F.~Gusfafsson, and T.~Karlsson.
\newblock Particle filtering and cramer-rao lower bound for underwater
  navigation.
\newblock In {\em Acoustics, Speech, and Signal Processing, 2003. Proceedings.
  (ICASSP '03). 2003 IEEE International Conference on}, volume~6, pages
  VI--65--8 vol.6, April 2003.

\bibitem{Silver2004}
D.~Silver, D.~Bradley, and S.~Thayer.
\newblock Scan matching for flooded subterranean voids.
\newblock In {\em IEEE Conference on Robotics, Automation and Mechatronics,
  2004.}, volume~1, pages 422--427 vol.1, Dec 2004.

\bibitem{Wirth2008}
Stephan Wirth, Alberto Ortiz, Dietrich Paulus, and Gabriel Oliver.
\newblock Using particle filters for autonomous underwater cable tracking*.
\newblock {\em IFAC Proceedings Volumes}, 41(1):161 -- 166, 2008.
\newblock 2nd IFAC Workshop on Navigation, Guidance and Control of Underwater
  Vehicles.

\bibitem{Maurelli2008}
F.~Maurelli, S.~Krupinski, Y.~Petillot, and J.~Salvi.
\newblock A particle filter approach for auv localization.
\newblock In {\em OCEANS 2008}, pages 1--7, 2008.

\bibitem{Maurelli2009}
F.~Maurelli, Y.~Petillot, A.~Mallios, P.~Ridao, and S.~Krupinski.
\newblock Sonar-based auv localization using an improved particle filter
  approach.
\newblock In {\em OCEANS 2009-EUROPE}, pages 1--9, May 2009.

\bibitem{Petillot2010}
Y.~Petillot, F.~Maurelli, N.~Valeyrie, A.~Mallios, P.~Ridao, J.~Aulinas, and
  J.~Salvi.
\newblock Acoustic-based techniques for auv localisation.
\newblock {\em Journal of Engineering for Maritime Environment},
  224(4):293--307, 2010.

\bibitem{Miguelanez:Semantic:09}
Emilio Miguelanez, Pedro Patr\'on, Keith Brown, Yvan~R. Petillot, and David~M.
  Lane.
\newblock Semantic knowledge-based framework to improve the situation awareness
  of autonomous underwater vehicles.
\newblock {\em IEEE Transactions on Knowledge and Data Engineering ({I}n
  {P}ress)}, PP(99), 2010.

\bibitem{pandora2012T}
David~M. Lane, Francesco Maurelli, Tom Larkworthy, Darwin Caldwell, Joaquim
  Salvi, Maria Fox, and Konstantinos Kyriakopoulos.
\newblock Pandora: Persistent autonomy through learning, adaptation,
  observation and re-planning.
\newblock {\em IFAC Proceedings Volumes}, 45(5):367 -- 372, 2012.
\newblock 3rd IFAC Workshop on Navigation, Guidance and Control of Underwater
  Vehicles.

\bibitem{pandora2016}
F.~Maurelli, M.~Carreras, J.~Salvi, D.~Lane, K.~Kyriakopoulos, G.~Karras,
  M.~Fox, D.~Long, P.~Kormushev, and D.~Caldwell.
\newblock The pandora project: A success story in auv autonomy.
\newblock In {\em OCEANS 2016 - Shanghai}, pages 1--8, April 2016.

\bibitem{Sinha2012}
Hyon Lim and Sudipta~N. Sinha.
\newblock Towards real-time semantic localization.
\newblock In {\em ICRA workshop on semantic perception}, 2012.

\bibitem{Martinez-Gomez2016}
Jesus Martínez-Gómez, Vicente Morell, Miguel Cazorla, and Ismael
  García-Varea.
\newblock Semantic localization in the pcl library.
\newblock {\em Robotics and Autonomous Systems}, 75:641 -- 648, 2016.

\bibitem{rubio2014dynamic}
Fernando Rubio, M~Julia Flores, Jes{\'u}s~Mart{\i}nez G{\'o}mez, and Ann
  Nicholson.
\newblock Dynamic bayesian networks for semantic localization in robotics.
\newblock In {\em XV Workshop of Physical Agents: Book of Proceedings, WAF
  2014, June 12th and 13th, 2014 Le{\'o}n, Spain}, pages 144--155, 2014.

\bibitem{Villena2015}
Álvaro Villena, Ismael García-Varea, Jesus Martínez-Gómez, Luis
  Rodríguez-Ruiz, and Cristina Romero-González.
\newblock A study of robot semantic localization based on multimodal hri, 11
  2015.

\bibitem{yi2009active}
Chuho Yi, Il~Hong Suh, Gi~Hyun Lim, and Byung-Uk Choi.
\newblock Active-semantic localization with a single consumer-grade camera.
\newblock In {\em Systems, Man and Cybernetics, 2009. SMC 2009. IEEE
  International Conference on}, pages 2161--2166. IEEE, 2009.

\bibitem{Morse}
G.~Echeverria, N.~Lassabe, A.~Degroote, and S.~Lemaignan.
\newblock Modular open robots simulation engine: Morse.
\newblock In {\em 2011 IEEE International Conference on Robotics and
  Automation}, pages 46--51, May 2011.

\end{thebibliography}
\end{document}